\documentclass[lettersize,journal]{IEEEtran}
\usepackage{amsmath,amsfonts}
\usepackage{algorithmic}
\usepackage{algorithm}
\usepackage{array}
\usepackage[caption=false,font=normalsize,labelfont=sf,textfont=sf]{subfig}
\usepackage{textcomp}
\usepackage{stfloats}
\usepackage{url}
\usepackage{verbatim}
\usepackage{graphicx}
\usepackage{cite}
\usepackage{multirow}
\usepackage{multicol}
\usepackage{makecell}
\usepackage{rotating}
\usepackage{booktabs}
\usepackage[table]{xcolor}
\usepackage[switch]{lineno}
\usepackage[colorlinks=true,citecolor=blue,linkcolor=blue,urlcolor=blue]{hyperref}

\newcommand{\pars}[1]{\noindent\textbf{#1}}
\newcommand{\nmc}[2]{#1 $\pm$ #2}
\newcommand{\bmc}[2]{\textbf{#1 $\pm$ #2}}
\newcommand{\smc}[2]{\underline{#1 $\pm$ #2}}
\newcommand{\tmc}[2]{\textit{#1 $\pm$ #2}}
\newcommand{\modeltab}{\textbf{ScaleBrain}}

\begin{document}

\title{ScaleBrain: A Scale-Consistent Framework for EEG Foundation Models}

\author{Minyu Zhou, Yajing Wu, Yongqiang Tang, Wang Jia and Wensheng Zhang
\thanks{This work was supported in part by the National Key Research and Development Program of China (No. 2021ZD0201600), in part by the National Natural Science Foundation of China under Grant 62476274 and Grant U22B2048, and in part by the Special Project of Joint Funding for Municipal, University (Institute) and Enterprise under the Guangzhou Basic Research Program (2024A03J0395), Science and Technology Projects in Guangzhou (2024D03J0010). (\textit{Corresponding authors: Yajing Wu and Yongqiang Tang.})}
\thanks{Minyu Zhou is with the College of Computer Science, Nankai University, Tianjin 300350, China (e-mail: minyu.zhou@mail.nankai.edu.cn).}%
\thanks{Yajing Wu and Yongqiang Tang are with the State Key Laboratory of Multimodal Artificial Intelligence Systems, Institute of Automation, Chinese Academy of Sciences, Beijing 100190, China (e-mail: yajing.wu@ia.ac.cn; yongqiang.tang@ia.ac.cn).}%
\thanks{Wang Jia is with the Department of Neurosurgery, Beijing Tiantan Hospital, Capital Medical University, Beijing 100070, China (e-mail: jwttyy@mail.ccmu.edu.cn).}%
\thanks{Wensheng Zhang is with the College of Computer Science, Nankai University, Tianjin 300350, China, and also with the School of Computer Science and Cyber Engineering, Guangzhou University, Guangzhou 510006, China (e-mail: zhangwenshengia@hotmail.com).}
}

\markboth{Journal of \LaTeX\ Class Files,~Vol.~14, No.~8, August~2021}%
{Zhou \MakeLowercase{\textit{et al.}}: ScaleBrain: A Scale-Consistent Framework for EEG Foundation Models}

\IEEEpubid{0000--0000/00\$00.00~\copyright~2021 IEEE}

\IEEEaftertitletext{\vspace{-1em}}
\maketitle

\nolinenumbers

\begin{abstract}
Foundation models pre-trained on large unlabeled electroencephalography (EEG) corpora have emerged as a transformative paradigm for Brain-Computer Interface (BCI) applications. 
Recent EEG foundation models have shown that multi-scale representations  derived from fixed-length patches can improve EEG representation learning.
Existing frameworks  commonly construct temporal scales through feature-level operations. However, these operations leave the correspondence between learned features and scale-specific characteristics of the original EEG signals implicit. Moreover, when the self-supervised objective is not aligned with the multi-scale input representations, information at some temporal scales may receive insufficient supervision during pre-training.
In light of this, we propose \emph{ScaleBrain}, a scale-consistent pre-training framework that explicitly coordinates multi-scale representations of EEG from both the model input and optimization objective sides. Specifically, ScaleBrain introduces \emph{multi-scale EEG tokenization} to represent the same EEG segment through analysis windows with different temporal spans and shared temporal anchors, and pairs these representations with \emph{scale-specific reconstruction} to align with the multi-scale input structure and provide supervision across temporal resolutions during self-supervised learning.
We further develop an axial spatio-temporal attention mechanism to contextualize the fused multi-scale representation along spatial and temporal dimensions.
Extensive evaluations on six public BCI benchmarks against 12 representative baselines show that ScaleBrain achieves the highest mean Balanced Accuracy on five datasets and ranks third on the remaining one, with only 5.0M parameters, demonstrating strong performance and favorable parameter efficiency.
\end{abstract}


\section{Introduction}

\begin{figure}[ht]
    \centering
    \includegraphics[width=\linewidth]{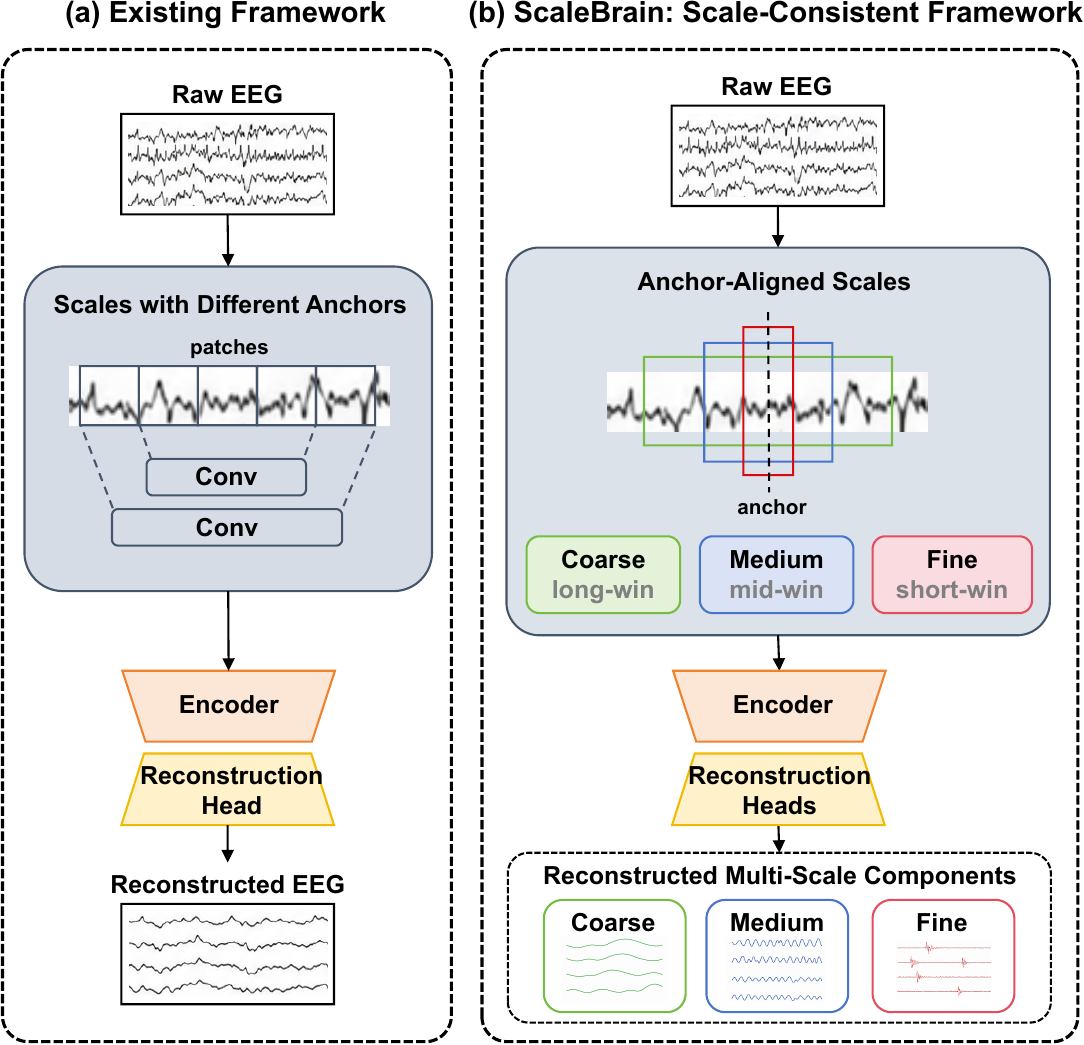}
    \caption{Comparison of pre-training paradigms for EEG foundation models. Unlike \textbf{(a)} the existing multi-scale framework which focuses on the input side with different temporal anchors, our proposed \textbf{(b)} ScaleBrain explicitly maintains scale consistency by coordinating coarse-to-fine representations across input construction and self-supervised optimization. Long/mid/short-win: long/middle/short window.}
    \label{fig:motivation}
\end{figure}

Understanding and interpreting brain activity has long been a central goal in cognitive neuroscience and artificial intelligence~\cite{kay2008identifying,palazzo2020decoding}. Scalp-based electroencephalography (EEG), which records electrical potentials at the scalp surface, is widely used as a non-invasive brain recording modality due to its high temporal resolution and practical feasibility~\cite{biasiucci2019electroencephalography}. By decoding multi-channel EEG signals, Brain-Computer Interface (BCI) technologies have enabled a wide range of applications, including motor imagery, emotion recognition, sleep staging, and abnormality detection~\cite{al2021motor,chen2026emod,wang2024sleep,roy2019chrononet}.

\IEEEpubidadjcol

Over the past few decades, EEG decoding methods have evolved through three major paradigms: traditional statistical methods~\cite{blankertz2007optimizing,guler2007multiclass}, task-specific deep learning models~\cite{lawhern2018eegnet,demir2021eeggnn,song2021transformer}, and more recently, EEG foundation models that learn transferable representations across diverse  tasks~\cite{yang2023biot,jiang2024labram,zhou2025csbrain}. This paradigm shift toward EEG foundation models follows the success of large-scale pre-training in natural language processing and computer vision, where Self-Supervised Learning (SSL) enables transferable representation learning from unlabeled data. Similarly, EEG foundation models leverage SSL objectives such as masked signal modeling~\cite{wang2025cbramod,zhou2025csbrain} and neural code prediction~\cite{jiang2024labram} to learn from large-scale unlabeled EEG corpora, capturing transferable patterns in EEG signals without requiring exhaustive manual annotations. Once pre-trained, these representations can be adapted to downstream BCI tasks through fine-tuning, especially when labeled data is limited.
However, the effectiveness of such transfer depends on whether the pre-trained representations adequately capture the diverse temporal characteristics of EEG signals.

Neural activity can exhibit both slowly varying rhythmic dynamics and rapidly changing transient patterns~\cite{buzsaki2004neuronal,canolty2010functional}. These dynamics may be represented at different temporal and spectral resolutions, motivating the use of multiple modeling scales in EEG analysis.
As illustrated in Fig.~\ref{fig:motivation}(a), recent EEG foundation models such as CSBrain~\cite{zhou2025csbrain} and CodeBrain~\cite{ma2026codebrain}, which build multi-scale representations based on patches and convolutions at the feature level, have demonstrated the potential of incorporating multi-scale mechanisms into large-scale EEG representation learning. 
Nevertheless, existing approaches commonly introduce temporal scales through feature-level operations, potentially overlooking scale-dependent temporal and spectral characteristics of the original EEG signals. The correspondence between the resulting representations and these signal characteristics remains implicit. Moreover, when the self-supervised objective is not aligned with the multi-scale representations, information at some temporal scales may receive insufficient supervision during pre-training.

\emph{On the one hand,} constructing multiple temporal scales introduces a correspondence issue. Applying an EEG tokenizer on the EEG signals with multiple window lengths produces tokens with different temporal spans and spectral resolutions. Although these tokens originate from the same EEG segment, their temporal semantics are not directly comparable unless their temporal anchors are explicitly aligned. Without such alignment, directly combining heterogeneous tokens may obscure the correspondence between representations at different temporal scales and make subsequent multi-scale fusion difficult.
\emph{On the other hand,} multi-scale design at the input side is insufficient if the self-supervised objective does not preserve the same scale structure. Existing EEG foundation models typically rely on a single-scale self-supervised objective, which provides supervision at only one temporal resolution. Such an objective may bias representation learning toward the characteristics of a particular scale and leaves information in other temporal scales insufficiently supervised, leading to potentially incomplete representations. Consequently, an effective multi-scale foundation model should therefore maintain consistency between scale diversity in the input representation and scale-specific supervision in the self-supervised objective. Meanwhile, representations at different scales should be jointly fused and contextualized as they are inherently coupled.

In this study, we propose \textbf{ScaleBrain}, a scale-consistent framework for EEG foundation models that explicitly coordinates multiple temporal scales during self-supervised pre-training, as illustrated in Fig.~\ref{fig:motivation}(b). Specifically, ScaleBrain uses \emph{multi-scale tokenization} to construct multiple representations of the same raw EEG segment using analysis windows with different temporal spans but shared temporal anchors, thereby establishing correspondence among tokens at different scales. At each scale, a spectral branch captures time-frequency information while a temporal branch captures waveform-level dynamics. The resulting representations are equipped with scale-aware spatio-temporal position encodings and integrated through gated scale fusion into a unified representation. The fused representation is then contextualized by an axial spatio-temporal attention mechanism along the spatial and temporal dimensions. Finally, ScaleBrain introduces \emph{scale-specific reconstruction} targets so that the scale structure introduced at the input side is explicitly preserved in the self-supervised objective. In this way, ScaleBrain explicitly models scale diversity from both the input and objective sides while maintaining a common temporal reference across scales.

The contributions are summarized as follows:
\begin{itemize}
    \item We formulate EEG foundation-model pre-training from the perspective of scale consistency, and propose ScaleBrain, which consistently models diverse temporal scales from both the input representation and self-supervised objective sides.
    \item We introduce a multi-scale EEG tokenization strategy that analyzes the same raw EEG segment using multiple temporal windows with shared temporal anchors. The resulting scale-specific representations preserve correspondence across temporal resolutions while capturing complementary spectral and temporal information.
    \item We introduce scale-specific reconstruction to provide objective-side supervision at multiple temporal scales aligned to input representations, and employ gated scale fusion and axial spatio-temporal attention to integrate and contextualize the multi-scale representations.
    \item We evaluate ScaleBrain on six public benchmark datasets spanning diverse downstream BCI tasks. The results show that ScaleBrain obtains the highest mean Balanced Accuracy on five datasets and the third-highest on the remaining one, while using only 5.0M parameters, validating the effectiveness of the proposed framework.
\end{itemize}

\section{Related Work}
\subsection{Task-Specific EEG Decoding}
Historically, EEG decoding relied heavily on hand-crafted features, such as power spectral density and common spatial patterns, combined with shallow classifiers like support vector machines~\cite{blankertz2007optimizing,guler2007multiclass}. With advances in deep learning, Convolutional Neural Networks (CNNs) such as EEGNet~\cite{lawhern2018eegnet} and DeepConvNet~\cite{schirrmeister2017conv} have become standard backbones for capturing spatio-temporal patterns. More recently, Transformer-based architectures, such as EEGConformer~\cite{song2022conformer}, have been introduced to model long-range temporal dependencies. While these task-specific models have achieved competitive performance on particular datasets, they are typically trained from scratch, require substantial labeled data, and often suffer from poor generalization due to high inter-subject variability and cross-session domain shifts.

\subsection{Self-Supervised EEG Foundation Models}
To mitigate the scarcity of labeled EEG data, recent research has shifted toward SSL to build EEG foundation models. Pre-trained on massive unlabeled EEG corpora, models like BIOT~\cite{yang2023biot}, LaBraM~\cite{jiang2024labram}, CBraMod~\cite{wang2025cbramod}, BraSTORM~\cite{wang2026brastorm}, and REVE~\cite{el2026reve} utilize masked autoencoding or contrastive objectives to learn generalizable neural representations.
Despite their impressive generalization across diverse downstream tasks, these models typically partition EEG signals into a single series of fixed-length patches. 
Recent multi-scale foundation models have alleviated this limitation by introducing representations at multiple temporal scales. However, how these scales should be temporally coordinated and consistently supervised during pre-training remains less explored.
Our proposed ScaleBrain departs from these methodologies by explicitly modeling multi-scale dynamics from both the input and objective sides to form a consistent multi-scale framework.

\subsection{Multi-Scale EEG Modeling}

Multi-scale modeling has been widely explored in EEG analysis. Traditional signal-processing methods often use band-pass filtering to extract frequency-specific features~\cite{ang2008filter}, while task-specific deep models employ temporal convolutions with multiple receptive fields~\cite{ding2022tsception}, multi-resolution spatio-temporal graph learning~\cite{fan2025medgnn}, or multi-granularity patching~\cite{wang2024medformer} to capture patterns at different temporal resolutions.
Within EEG foundation modeling, CSBrain~\cite{zhou2025csbrain} and CodeBrain~\cite{ma2026codebrain} explicitly incorporate multi-scale representations, demonstrating the potential of modeling resolution diversity during pre-training. Nevertheless, the use of multiple scales does not by itself ensure consistent correspondence and supervision across the pre-training pipeline. In contrast, ScaleBrain treats scale consistency as a property that should be maintained across both the input and objective levels: different analysis windows are temporally aligned through shared anchors, and each scale is associated with a dedicated reconstruction target. This formulation allows scale information and temporal correspondence to be preserved from tokenization through self-supervised optimization.

\section{Methodology}

\begin{figure*}[ht]
    \centering
    \includegraphics[width=\linewidth]{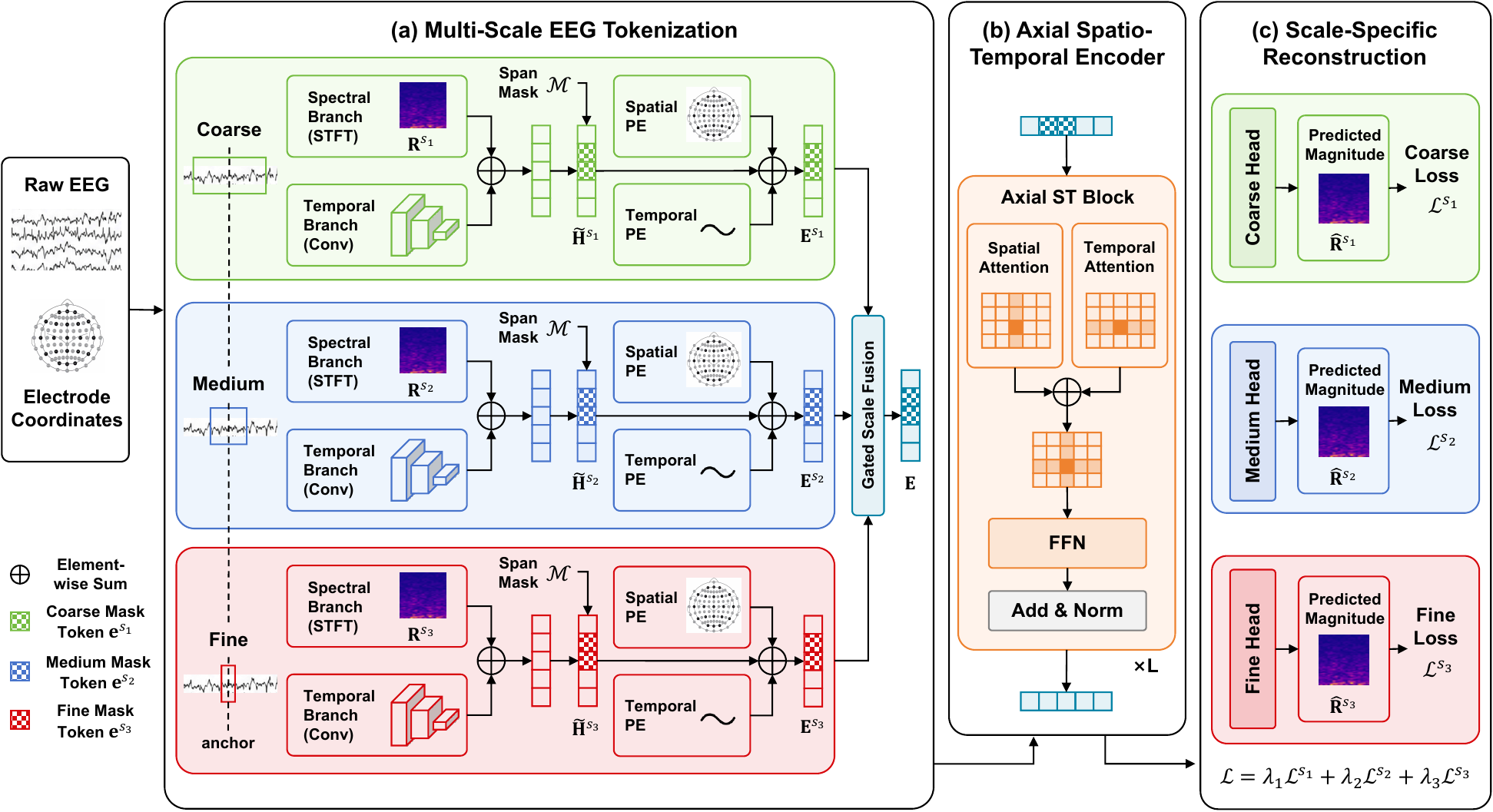}
    \caption{The overall framework of ScaleBrain, which is composed of \textbf{(a)} multi-scale EEG tokenization, \textbf{(b)} axial spatio-temporal encoder, and \textbf{(c)} scale-specific reconstruction. PE: Positional Encoding. FFN: Feed-Forward Network.}
    \label{fig:framework}
\end{figure*}

\subsection{Framework Overview and Notations}

Fig.~\ref{fig:framework} illustrates the overall framework of ScaleBrain, which reformulates EEG foundation-model pre-training as a scale-consistent masked reconstruction problem, addressing both the need for temporal scale correspondence at the input side and the need for scale-consistent supervision at the objective side. Given an EEG segment, the framework  extracts coarse-, medium-, and fine-scale representations using temporally aligned spectral and temporal branches. These scale-specific representations are then masked by a shared span mask, enriched with scale-aware spatio-temporal position encodings, fused into a unified representation, encoded by an axial spatio-temporal Transformer, and finally used to reconstruct masked spectral targets at each scale.

Formally, let $\mathbf{X}\in\mathbb{R}^{C\times T}$ denote a preprocessed EEG segment, where $\mathcal{C}_0$ is a reference electrode set with known coordinates following the international 10-10 system, $\mathcal{C}\subseteq\mathcal{C}_0$ is the observed electrode set, $C=|\mathcal{C}|$ is the number of channels, and $T$ is the number of time points. ScaleBrain defines EEG scales by the temporal span of the analysis window, where $s_n$ denotes the number of sampled timepoints at the $n$-th scale and corresponds to $s_n/f_s$ seconds under sampling rate $f_s$. The pre-training is performed over a set of temporal scales $\mathcal{S}=\{s_1,\ldots,s_N\}$. For each scale $s_n$, the model produces a scale-specific representation $\mathbf{E}^{s_n}\in\mathbb{R}^{C\times M\times D}$, where $M$ is the number of temporally aligned windows and $D$ is the hidden dimension. The representations from all scales are aligned and fused into $\mathbf{E}\in\mathbb{R}^{C\times M\times D}$, which is then processed by the axial spatio-temporal encoder.

\subsection{Multi-Scale EEG Tokenization}

To capture the complementary temporal information represented at different scales, we propose a multi-scale EEG tokenization that does not rely on a single series of fixed-length patches but generates temporally aligned tokens across multiple window scales, as presented in Fig.~\ref{fig:framework}(a).
Longer windows provide coarser temporal views but finer frequency resolution, making them suitable for characterizing slow rhythmic structures, whereas shorter windows provide finer temporal localization at the cost of coarser frequency resolution, helping preserve transient and rapidly varying patterns. We use a shared hop size $h$ and define the window anchor locations as $\tau_i=ih$, where $0\le i\le M-1$ denotes the window index, ensuring that tokens from different scales share a common temporal reference. The windows are centred at $\tau_i$ with zero padding at segment boundaries, so $M$ is identical across scales. This explicit temporal correspondence provides a common basis for cross-scale fusion and representation learning.

\pars{Spectral Branch.}
For each scale $s_n$, the spectral branch applies Short-Time Fourier Transform (STFT) with window length $s_n$ and a shared hop size $h$. Given an input EEG segment $\mathbf{X}$, it transforms the $c$-th channel $\mathbf{X}_c$ into a scale-specific spectrogram $\mathbf{Z}^{s_n}_c\in\mathbb{C}^{M \times F^{s_n}}$:
\begin{equation}
    \mathbf{Z}_c^{s_n}(\tau_i)
    = \mathcal{F}\!\big[\mathbf{X}_c(t) w^{s_n}(t-\tau_i)\big],
    \quad 1 \le c \le C,
\end{equation}
where $t$ is time, $w^{s_n}(\cdot)$ is the windowing function for scale $s_n$, $\mathcal{F}[\cdot]$ denotes the Fourier transform, and $F^{s_n}$ is the number of frequency bins at this scale.
The channel-wise spectrograms are stacked to form $\mathbf{Z}^{s_n}\in\mathbb{C}^{C\times M\times F^{s_n}}$. We use the log magnitude $\mathbf{R}^{s_n}\in\mathbb{R}^{C\times M\times F^{s_n}}$ computed from the EEG segment as the scale-specific reconstruction target:
\begin{equation}
    \mathbf{R}^{s_n}=\log\!\big(1+|\mathbf{Z}^{s_n}|\big).
\end{equation}
The spectral feature $\mathbf{H}_\text{spec}^{s_n}\in \mathbb{R}^{C\times M\times D}$ is obtained by projecting this target into the hidden space with a scale-specific projector transforming the feature dimension:
\begin{equation}
    \mathbf{H}_\text{spec}^{s_n}
    = \operatorname{Proj}^{s_n}(\mathbf{R}^{s_n}).
\end{equation}

\pars{Temporal Branch.} To complement the spectral representation with waveform-level temporal details, we introduce a temporal branch in parallel with the spectral branch. Specifically, we use a three-layer convolutional network, which uses kernel sizes matched to $s_n$ and strides/paddings matched to $h$, so that its receptive field and temporal hop are aligned with the spectral tokens at the same scale. The temporal branch produces the temporal feature $\mathbf{H}_\text{temp}^{s_n}\in\mathbb{R}^{C\times M\times D}$.

\pars{Scale-Aware Spatio-Temporal Position Encodings.} We propose scale-aware spatio-temporal Position Encodings (PEs) to guide the Transformer in distinguishing scale-specific representations while preserving their common spatial and temporal correspondence. Unlike conventional models that share a universal positional encoding, our scale-aware PEs project spatial coordinates and temporal bases into scale-specific representation spaces, allowing the model to distinguish different temporal granularities across scales.
For scale $s_n$, the scale-aware spatial positional encoding $\mathbf{E}_\text{spat}^{s_n}\in\mathbb{R}^{C\times D}$ is obtained by a two-layer multi-layer perceptron from the electrode coordinates, while the scale-aware temporal positional encoding $\mathbf{E}_\text{temp}^{s_n}\in\mathbb{R}^{M\times D}$ is built upon a fixed sinusoidal basis, with a projector injecting scale-specific temporal information.
Notably, for bipolar channels, the coordinate is defined as the midpoint of the two constituent electrodes in the 10-10 layout, which keeps the spatial PE well-defined while preserving the geometry of the montage.

To enable masked self-supervised pre-training, we apply a scale-aware span masking strategy to the tokens. We randomly generate a shared span mask $\mathcal{M}\subset\{(c,m)\mid 1\le c\le C,1\le m\le M\}$ with a mask ratio $r$ and a span length $l$. Here, $l$ means the number of consecutive tokens grouped as one masking unit along the temporal dimension, and the units are selected so that approximately $r$ of the tokens are masked.
For each scale $s_n$, the tokens at the masked positions are replaced with a learnable scale-specific mask token $\mathbf{e}^{s_n}\in\mathbb{R}^{D}$:
\begin{equation}
\tilde{\mathbf{H}}^{s_n}=\operatorname{SpanMask}\!\left(\mathbf{H}_\text{spec}^{s_n}+\mathbf{H}_\text{temp}^{s_n}, \mathbf{e}^{s_n}\right).
\end{equation}
The scale-specific token embeddings are then formed by adding the spatial and temporal PEs with dimension broadcasting, followed by RMSNorm~\cite{zhang2019root}:
\begin{equation}
    \mathbf{E}^{s_n}=\operatorname{RMSNorm}\!\left(\tilde{\mathbf{H}}^{s_n}+\mathbf{E}_\text{spat}^{s_n}+\mathbf{E}_\text{temp}^{s_n}\right).
\end{equation}

\pars{Gated Scale Fusion.} We leverage a gated projection to fuse multiple scale embeddings into a unified representation. The gated projection is inspired by SwiGLU~\cite{shazeer2020glu}:
\begin{equation}
    \operatorname{GatedProj}(\mathbf{x})=\operatorname{Swish}(\mathbf{x}\mathbf{W}^g)\odot \mathbf{x}\mathbf{W}^o,
\end{equation}
where $\mathbf{W}^o$ and $\mathbf{W}^g$ are learnable projection matrices, and $\odot$ denotes the Hadamard product. Here, $\mathbf{x}\mathbf{W}^o$ produces the projected scale-fusion output, while $\operatorname{Swish}(\mathbf{x}\mathbf{W}^g)$ acts as a data-dependent gate that modulates this output. Unlike the original SwiGLU used mainly within feed-forward layers, our gated projection is used to adaptively integrate complementary representations from different temporal scales into a unified representation. The fused embedding $\mathbf{E}$ is produced by projecting the concatenation of the scale-specific embeddings:
\begin{equation}
    \mathbf{E}=\operatorname{GatedProj}\!\big([\mathbf{E}^{s_1};\cdots;\mathbf{E}^{s_N}]\big),
\end{equation}
where $[;]$ means concatenation along the feature dimension.
The fused representation $\mathbf{E}$ is then fed into the axial spatio-temporal encoder for contextualized token modeling.

\subsection{Axial Spatio-Temporal Encoder}

The Axial Spatio-Temporal (Axial ST) encoder takes the fused representation $\mathbf{E}$ as input and contextualizes it to model spatial and temporal dependencies in EEG. As shown in Fig.~\ref{fig:framework}(b), the encoder consists of $L$ stacked Axial ST Transformer blocks. Each Axial ST block contains Axial ST attention, a Feed-Forward Network (FFN), residual connections, and RMSNorm~\cite{zhang2019root}. The FFN uses SwiGLU~\cite{shazeer2020glu} to enhance nonlinear modeling capacity.

\pars{Axial ST Attention.}
The Axial ST attention consists of two parallel attentions: spatial attention and temporal attention, each using the full hidden dimension $D$ with $K$ attention heads.
The spatial attention $\operatorname{S-Attn}(\cdot)$ performs standard multi-head self-attention $\operatorname{MHA}(\cdot)$ along the spatial dimension at each temporal window:
\begin{equation}
\begin{aligned}
    &\mathbf{F}_m=\operatorname{MHA}\!\left(\mathbf{E}_m\mathbf{W}^Q_\text{spat},\mathbf{E}_m\mathbf{W}^K_\text{spat},\mathbf{E}_m\mathbf{W}^V_\text{spat}\right),\\
    &\operatorname{S-Attn}(\mathbf{E})=[\mathbf{F}_1,\cdots,\mathbf{F}_M],
\end{aligned}
\end{equation}
where $\mathbf{E}_m=\mathbf{E}_{:,m,:}\in\mathbb{R}^{C\times D}$, $1\le m\le M$ denotes the spatial-wise slice at the $m$-th temporal window, $\mathbf{W}^Q_\text{spat}$, $\mathbf{W}^K_\text{spat}$, $\mathbf{W}^V_\text{spat}$ are projection matrices of the query, key, and value in the spatial attention, and $[,]$ denotes stacking along the temporal dimension.
Analogously, the temporal branch applies multi-head self-attention across temporal windows for each channel, using $\mathbf{E}_c=\mathbf{E}_{c,:,:}\in\mathbb{R}^{M\times D},1\le c\le C$. The Axial ST attention output is obtained by summing the two branch outputs. 
Thus, the axial design further integrates the fused information across spatial and temporal dimensions while retaining the interactions among multiple temporal scales.

\subsection{Scale-Specific Reconstruction}

We pre-train ScaleBrain with scale-specific reconstruction targets illustrated in Fig.~\ref{fig:framework}(c), building upon the standard masked autoencoding framework.
Consistent with the multi-scale EEG tokenization, we use the scale-specific log-magnitude spectra $\mathbf{R}^{s_n}$ extracted by the spectral branches as reconstruction targets. Reconstructing these targets in parallel encourages the model to preserve information associated with each temporal scale throughout self-supervised learning.

For each scale $s_n$, we define the reconstruction loss $\mathcal{L}^{s_n}$ using  Mean Squared Error (MSE):
\begin{equation}
    \mathcal{L}^{s_n}=\frac{1}{F^{s_n}|\mathcal{M}|}\sum_{(c,m)\in\mathcal{M}}
    \left\|
    \hat{\mathbf{R}}^{s_n}_{c,m,:}-\mathbf{R}^{s_n}_{c,m,:}
    \right\|_2^2,
\end{equation}
where $\hat{\mathbf{R}}^{s_n}$ is the reconstructed log magnitudes for scale $s_n$, and $\mathcal{M}$ is the span mask shared across scales.
The overall pre-training objective $\mathcal{L}$ is a weighted sum of the scale-specific losses:
\begin{equation}
    \mathcal{L}=\sum_{n=1}^{N}\lambda_n \mathcal{L}^{s_n},
\end{equation}
where $\lambda_n$ is the weight for scale $s_n$.
The purpose of the scale-specific objective is to balance supervision across temporal resolutions rather than to explicitly separate frequency bands. Under equal weighting, the optimization may favor the coarse reconstruction scale which has simpler signal structure. We therefore assign progressively larger weights to finer temporal scales, reducing optimization bias toward the coarse-scale reconstruction target and encouraging the encoder to retain localized information.

\begin{table*}[ht]
    \centering
    \small
    \caption{Overview of downstream BCI tasks and datasets.}
    \begin{tabular}{llcccccc}
        \toprule
        \textbf{Task} & \textbf{Dataset} & \textbf{Rate (Hz)} & \textbf{Duration (s)} & \textbf{\#Channels} & \textbf{\#Samples} & \textbf{\#Subjects} & \textbf{Label} \\
        \midrule
        Motor Imagery             & PhysioNet-MI     & 160  & 4  & 64 & 9837   & 109  & 4-class \\
        Emotion Recognition       & SEED-V           & 1000 & 1  & 62 & 117744 & 16   & 5-class \\
        Sleep Staging             & HMC              & 256  & 30 & 4  & 137243 & 151  & 5-class \\
        Mental Stress Detection   & MentalArithmetic & 500  & 5  & 20 & 1707   & 36   & 2-class \\
        Abnormality Detection     & TUAB             & 250  & 10 & 16 & 409455 & 2383 & 2-class \\
        Event Type Classification & TUEV             & 250  & 5  & 16 & 112491 & 370  & 6-class \\
        \bottomrule
    \end{tabular}
    \label{tab:datasets}
\end{table*}

\section{Experiments}

\subsection{Pre-Training and Downstream Datasets}

\pars{Pre-Training Dataset Preprocessing.}
We pre-train ScaleBrain on the Temple University Hospital EEG corpus (TUEG)~\cite{obeid2016temple}, which is commonly used in EEG foundation model studies~\cite{wang2025cbramod,zhou2025csbrain}. Following prior preprocessing protocols~\cite{jiang2024labram,wang2025cbramod}, all signals are filtered by a 0.3-75 Hz band-pass filter and a 60 Hz notch filter, resampled to 200 Hz, and segmented into non-overlapping 30-second EEG samples. This preprocessing yields a total of 1,175,693 samples, corresponding to approximately 9,797 hours of EEG data. For consistency with prior foundation-model benchmarks~\cite{wang2025cbramod,zhou2025csbrain,ma2026codebrain}, our main comparison follows the commonly used full-TUEG pre-training protocol, in which recordings belonging to TUAB and TUEV datasets are retained in the pre-training corpus.

\pars{Downstream BCI Tasks and Datasets.}
We conduct evaluations across 6 public EEG datasets spanning various representative BCI tasks, as summarized in Table~\ref{tab:datasets}. These tasks include \textit{Motor Imagery Classification} (PhysioNet-MI~\cite{goldberger2000physionet}), \textit{Emotion Recognition} (SEED-V~\cite{liu2021emotion}), \textit{Sleep Staging} (HMC~\cite{alvarez2021sleep}), \textit{Mental Stress Detection} (MentalArithmetic~\cite{goldberger2000physionet,zyma2019arithmetic}), \textit{Abnormality Detection} (TUAB~\cite{obeid2016temple}), and \textit{Event Type Classification} (TUEV~\cite{obeid2016temple}). We detail the preprocessing and data split of each dataset in the following.

\begin{itemize}
    \item \textbf{PhysioNet-MI}~\cite{goldberger2000physionet} is a widely used Motor Imagery (MI) dataset collected from 109 subjects, with four MI tasks: left fist, right fist, both fists, and both feet. The recordings were collected at 160 Hz with 64 electrodes. Following prior work~\cite{zhou2025csbrain}, the signals are segmented into 4-second samples and resampled to 200 Hz. We follow the subject-independent split in prior work~\cite{zhou2025csbrain}, where subjects 1-70 are used for training, 71-89 for validation, and 90-109 for testing.
    \item \textbf{SEED-V}~\cite{liu2021emotion} is an emotion recognition dataset comprising 16 subjects, covering five categories of emotion (happy, sad, neutral, disgust, fear). The recordings were collected at 1000 Hz with 62 electrodes. Following prior work~\cite{zhou2025csbrain}, we segment the signals into 1-second samples and resample to 200 Hz, and divide the 15 trials of each subject into three equal parts for training, validation, and testing.
    \item \textbf{HMC}~\cite{alvarez2021sleep} is a large-scale sleep dataset containing 151 subjects, annotated with five standard sleep stages (Wake, N1, N2, N3, REM). The recordings were collected at 256 Hz with 4 electrodes. Following prior work~\cite{zhou2025csbrain}, the signals are segmented into 30-second samples and resampled to 200 Hz. The data is split in a subject-independent manner, where subjects 1-100 are used for training, 101-125 for validation, and 126-151 for testing.
    \item \textbf{MentalArithmetic}~\cite{goldberger2000physionet,zyma2019arithmetic} is a small-scale dataset intended to identify mental stress, where recordings collected before the mental arithmetic task are labeled as ``without mental stress'', and recordings collected during the task are labeled as ``with mental stress''. The recordings were collected at 500 Hz with 20 electrodes. Following prior work~\cite{zhou2025csbrain}, the signals are segmented into 5-second samples and resampled to 200 Hz. We adopt a subject-independent split as in prior work~\cite{zhou2025csbrain}, where subjects 1-28 are used for training, 29-32 for validation, and 33-36 for testing.
    \item \textbf{TUAB}~\cite{obeid2016temple} is a subset of the TUEG corpus that focuses on the abnormality detection (or ``abnormal detection'' in some literature) task, with annotations of ``normal'' or ``abnormal''. The recordings were collected at 250 Hz with 23 electrodes. The signals are segmented into 10-second samples, resampled to 200 Hz, and converted to the commonly adopted 16-channel bipolar montage as in prior works~\cite{wang2025cbramod,zhou2025csbrain}. The training and testing subjects are specified by the original dataset. Following prior work~\cite{zhou2025csbrain}, we split the training subjects into training and validation sets at a 8:2 ratio.
    \item \textbf{TUEV}~\cite{obeid2016temple} is a subset of the TUEG corpus that focuses on event type classification, containing six types of events (SPSW, GPED, PLED, EYEM, ARTF, BCKG). The recordings were collected at 250 Hz with 23 electrodes. The signals are segmented into 5-second samples, resampled to 200 Hz, and converted to the commonly adopted 16-channel bipolar montage as in prior works~\cite{wang2025cbramod,zhou2025csbrain}. The training and testing subjects are specified by the original dataset. Following prior work~\cite{zhou2025csbrain}, we split the training subjects into training and validation sets at a 8:2 ratio.
\end{itemize}

\subsection{Implementation Details}

\pars{Pre-Training Settings.}
To instantiate the coarse-to-fine temporal scales defined in our tokenization module, we use $N=3$ scales, with $s_1=200$, $s_2=100$, and $s_3=40$. The shared hop size is set to $h=20$.  We use a masking ratio of $r=50\%$ and a span length of $l=10$ tokens. The objective weights are fixed as $\lambda_1=0.5$, $\lambda_2=1.0$, and $\lambda_3=2.0$, matching their design purpose. The encoder contains $L=8$ Axial ST blocks with hidden dimension $D=192$. We pre-train ScaleBrain for 5 epochs with a batch size of 128. Optimization is performed using AdamW with a learning rate of $2\times10^{-4}$, weight decay of $1\times10^{-2}$, and a cosine annealing learning-rate schedule with warmup. The warmup is a linear incremental process from zero to the target learning rate in the 10\% of total steps, then the cosine annealing takes over.
The detailed pre-training hyperparameters are listed in Table~\ref{tab:pretrain_hyperparams}, which are fixed before evaluation and are not tuned for any individual downstream datasets. Pre-training is conducted on a Linux server equipped with four NVIDIA RTX 3090 GPUs, each with 24 GB VRAM, using Python 3.12 and PyTorch 2.7.1 with CUDA 12.8.

\pars{Fine-Tuning Settings.}
To fine-tune ScaleBrain on a downstream dataset, we load the pre-trained weights and replace the reconstruction heads with a classification head composed of temporal or spatial pooling and a two-layer perceptron. Specifically, we apply spatial pooling to HMC and TUEV, and temporal pooling to other datasets. Because the model does not use channel-specific parameters tied to a fixed electrode set and instead encodes channel geometry through electrode coordinates, the pre-trained encoder can be applied to downstream datasets with different numbers of channels.
We then fine-tune the whole model on the target dataset with default hyperparameters listed in Table~\ref{tab:downstream_hyperparams}. We use binary cross-entropy loss for binary classification and cross-entropy for multi-class classification.

\begin{table}[t]
    \centering
    \small
    \caption{Pre-training hyperparameters for ScaleBrain.}
    \begin{tabular}{lc}
        \toprule
        \textbf{Component} & \textbf{Setting} \\
        \midrule
        Input Channels & 19 \\
        Time Points per Sample & 6000 \\
        Mask Ratio & 0.5 \\
        Mask Token & Learnable vector \\
        \midrule
        Number of Layers & 8 \\
        Hidden Dimension & 192 \\
        Attention Heads & 6 \\
        Feed-forward Dimension & 512 \\
        Dropout & 0.1 \\
        Weight Initialization & Kaiming Normal \\
        \midrule
        Optimizer & AdamW \\
        Learning Rate & 2e-4 \\
        Adam $\beta$ & (0.9, 0.999) \\
        Adam $\epsilon$ & 1e-8 \\
        Weight Decay & 1e-2 \\
        Gradient Clipping & 1.0 \\
        \midrule
        Training Epochs & 5 \\
        Batch Size & 128 \\
        Learning Rate Scheduler & CosineAnnealingLR \\
        Warmup Epochs & 0.5 \\
        Minimum Learning Rate & 1e-5 \\
        \bottomrule
    \end{tabular}
    \label{tab:pretrain_hyperparams}
\end{table}

\begin{table}[t]
    \centering
    \small
    \caption{Default hyperparameter settings for downstream fine-tuning.}
    \begin{tabular}{lc}
        \toprule
        \textbf{Component} & \textbf{Setting} \\
        \midrule
        Training Epochs & 10 \\
        Batch Size & 128 \\
        Optimizer & AdamW \\
        Learning Rate & 1e-4 \\
        Adam $\beta$ & (0.9, 0.999) \\
        Adam $\epsilon$ & 1e-8 \\
        Weight Decay & 1e-2 \\
        Dropout & 0.1 \\
        Learning Rate Scheduler & CosineAnnealingLR \\
        Warmup Epochs & 1 \\
        Minimum Learning Rate & 1e-6 \\
        Label Smoothing (for multi-class) & 0.1 \\
        \bottomrule
    \end{tabular}
    \label{tab:downstream_hyperparams}
\end{table}

\begin{table*}[ht]
    \centering
    \small
    \setlength{\tabcolsep}{0.3em}
    \caption{Comparison results of different methods across six EEG downstream tasks. The \textbf{best}, \underline{second-best}, and \textit{third-best} results are marked. * means recordings from TUAB and TUEV are excluded in pre-training. Acc.: Accuracy.}
    \begin{tabular}{lc|ccc|ccc}
        \toprule
        \textbf{Method} & \textbf{Params} & \multicolumn{3}{c|}{\textbf{PhysioNet-MI (4-class)}} & \multicolumn{3}{c}{\textbf{SEED-V (5-class)}} \\
        & & Balanced Acc. & Cohen's Kappa & Weighted F1 & Balanced Acc. & Cohen's Kappa & Weighted F1 \\
        \midrule
        EEGNet          & 0.003M & \nmc{0.5814}{0.0125} & \nmc{0.4468}{0.0199} & \nmc{0.5796}{0.0115} & \nmc{0.2961}{0.0102} & \nmc{0.1006}{0.0143} & \nmc{0.2749}{0.0098} \\
        EEGConformer    & 0.55M  & \nmc{0.6049}{0.0104} & \nmc{0.4736}{0.0171} & \nmc{0.6062}{0.0095} & \nmc{0.3537}{0.0112} & \nmc{0.1772}{0.0174} & \nmc{0.3487}{0.0136} \\
        SPaRCNet        & 0.76M  & \nmc{0.5932}{0.0152} & \nmc{0.4564}{0.0234} & \nmc{0.5937}{0.0147} & \nmc{0.2949}{0.0078} & \nmc{0.1121}{0.0139} & \nmc{0.2979}{0.0083} \\
        ContraWR        & 1.6M   & \nmc{0.5892}{0.0133} & \nmc{0.4527}{0.0248} & \nmc{0.5918}{0.0116} & \nmc{0.3546}{0.0105} & \nmc{0.1905}{0.0188} & \nmc{0.3544}{0.0121} \\
        FFCL            & 2.4M   & \nmc{0.5726}{0.0092} & \nmc{0.4323}{0.0182} & \nmc{0.5701}{0.0079} & \nmc{0.3641}{0.0092} & \nmc{0.2078}{0.0201} & \nmc{0.3645}{0.0132} \\
        ST-Transformer  & 3.5M   & \nmc{0.6035}{0.0081} & \nmc{0.4712}{0.0199} & \nmc{0.6053}{0.0075} & \nmc{0.3052}{0.0072} & \nmc{0.1083}{0.0121} & \nmc{0.2833}{0.0105} \\
        \midrule
        BIOT            & 3.2M   & \nmc{0.6153}{0.0154} & \nmc{0.4875}{0.0272} & \nmc{0.6158}{0.0197} & \nmc{0.3837}{0.0187} & \nmc{0.2261}{0.0262} & \nmc{0.3856}{0.0203} \\
        LaBraM-Base     & 5.8M   & \nmc{0.6173}{0.0122} & \nmc{0.4912}{0.0192} & \nmc{0.6177}{0.0141} & \nmc{0.3976}{0.0138} & \nmc{0.2386}{0.0209} & \nmc{0.3974}{0.0111} \\
        CBraMod         & 4.0M   & \nmc{0.6174}{0.0036} & \nmc{0.4898}{0.0048} & \nmc{0.6179}{0.0035} & \nmc{0.4091}{0.0097} & \nmc{0.2569}{0.0143} & \nmc{0.4101}{0.0108} \\
        CSBrain         & 4.9M   & \tmc{0.6304}{0.0090} & \tmc{0.5071}{0.0120} & \tmc{0.6308}{0.0095} & \bmc{0.4197}{0.0033} & \bmc{0.2785}{0.0034} & \bmc{0.4280}{0.0023} \\
        REVE-Base       & 69M    & \smc{0.6480}{0.0140} & \smc{0.5306}{0.0187} & \smc{0.6484}{0.0170} & \nmc{0.4048}{0.0023} & \nmc{0.2622}{0.0021} & \nmc{0.4145}{0.0017} \\
        CodeBrain & 15M       & -                    & -                    & -                          & \smc{0.4137}{0.0023} & \smc{0.2735}{0.0032} & \smc{0.4235}{0.0022} \\
        \modeltab       & 5.0M   & \bmc{0.6524}{0.0111} & \bmc{0.5365}{0.0148} & \bmc{0.6514}{0.0105} & \tmc{0.4116}{0.0056} & \tmc{0.2666}{0.0084} & \tmc{0.4181}{0.0076} \\
        \midrule
        \textbf{Method} & \textbf{Params} & \multicolumn{3}{c|}{\textbf{HMC (5-class)}} & \multicolumn{3}{c}{\textbf{MentalArithmetic (2-class)}} \\
        & & Balanced Acc. & Cohen's Kappa & Weighted F1 & Balanced Acc. & AUC-PR & AUROC \\
        \midrule
        EEGNet          & 0.003M & \nmc{0.6534}{0.0122} & \nmc{0.5886}{0.0201} & \nmc{0.6536}{0.0168} & \nmc{0.6770}{0.0116} & \nmc{0.5763}{0.0102} & \nmc{0.7321}{0.0108} \\
        EEGConformer    & 0.55M  & \nmc{0.7149}{0.0086} & \nmc{0.6432}{0.0055} & \nmc{0.7080}{0.0039} & \nmc{0.6805}{0.0123} & \nmc{0.5829}{0.0134} & \nmc{0.7424}{0.0128} \\
        SPaRCNet        & 0.76M  & \nmc{0.4756}{0.1109} & \nmc{0.3147}{0.1315} & \nmc{0.4108}{0.1310} & \nmc{0.6879}{0.0107} & \nmc{0.5825}{0.0193} & \nmc{0.7418}{0.0132} \\
        ContraWR        & 1.6M   & \nmc{0.4242}{0.0541} & \nmc{0.2340}{0.0554} & \nmc{0.2987}{0.0288} & \nmc{0.6631}{0.0097} & \nmc{0.5787}{0.0164} & \nmc{0.7332}{0.0082} \\
        FFCL            & 2.4M   & \nmc{0.4427}{0.0702} & \nmc{0.2542}{0.0654} & \nmc{0.2902}{0.0485} & \nmc{0.6798}{0.0142} & \nmc{0.5786}{0.0266} & \nmc{0.7330}{0.0198} \\
        ST-Transformer  & 3.5M   & \nmc{0.2559}{0.0141} & \nmc{0.0503}{0.0183} & \nmc{0.1428}{0.0122} & \nmc{0.6631}{0.0173} & \nmc{0.5672}{0.0259} & \nmc{0.7132}{0.0174} \\
        \midrule
        BIOT            & 3.2M   & \nmc{0.6862}{0.0041} & \nmc{0.6295}{0.0113} & \nmc{0.7091}{0.0147} & \nmc{0.6875}{0.0186} & \nmc{0.6004}{0.0195} & \nmc{0.7536}{0.0144} \\
        LaBraM-Base     & 5.8M   & \nmc{0.7277}{0.0101} & \nmc{0.6813}{0.0053} & \nmc{0.7454}{0.0027} & \nmc{0.6909}{0.0125} & \nmc{0.5999}{0.0155} & \nmc{0.7721}{0.0093} \\
        CBraMod         & 4.0M   & \nmc{0.7269}{0.0041} & \nmc{0.6685}{0.0104} & \nmc{0.7395}{0.0089} & \nmc{0.7256}{0.0132} & \nmc{0.6267}{0.0099} & \nmc{0.7905}{0.0073} \\
        CSBrain         & 4.9M   & \tmc{0.7345}{0.0047} & \tmc{0.6818}{0.0046} & \tmc{0.7506}{0.0042} & \tmc{0.7558}{0.0106} & \nmc{0.6696}{0.0221} & \tmc{0.8478}{0.0297} \\
        REVE-Base       & 69M    & \smc{0.7401}{0.0075} & \bmc{0.6982}{0.0078} & \smc{0.7638}{0.0074} & \smc{0.7660}{0.0355} & \smc{0.7470}{0.0807} & \nmc{0.8450}{0.0514} \\
        CodeBrain & 15M    & -                    & -                    & -                          & \nmc{0.7514}{0.0203} & \tmc{0.7177}{0.0421} & \smc{0.8707}{0.0209} \\
        \modeltab       & 5.0M   & \bmc{0.7540}{0.0098} & \smc{0.6961}{0.0145} & \bmc{0.7646}{0.0122} & \bmc{0.7771}{0.0429} & \bmc{0.8062}{0.0809} & \bmc{0.8872}{0.0437} \\
        \midrule
        \textbf{Method} & \textbf{Params} & \multicolumn{3}{c|}{\textbf{TUAB (2-class)}} & \multicolumn{3}{c}{\textbf{TUEV (6-class)}} \\
        & & Balanced Acc. & AUC-PR & AUROC & Balanced Acc. & Cohen's Kappa & Weighted F1 \\
        \midrule
        EEGNet          & 0.003M & \nmc{0.7642}{0.0036} & \nmc{0.8299}{0.0043} & \nmc{0.8412}{0.0031} & \nmc{0.3876}{0.0143} & \nmc{0.3577}{0.0155} & \nmc{0.6539}{0.0120} \\
        EEGConformer    & 0.55M  & \nmc{0.7758}{0.0049} & \nmc{0.8427}{0.0054} & \nmc{0.8445}{0.0038} & \nmc{0.4074}{0.0164} & \nmc{0.3967}{0.0195} & \nmc{0.6983}{0.0152} \\
        SPaRCNet        & 0.76M  & \nmc{0.7896}{0.0018} & \nmc{0.8414}{0.0018} & \nmc{0.8676}{0.0012} & \nmc{0.4161}{0.0262} & \nmc{0.4233}{0.0181} & \nmc{0.7024}{0.0104} \\
        ContraWR        & 1.6M   & \nmc{0.7746}{0.0041} & \nmc{0.8421}{0.0104} & \nmc{0.8456}{0.0074} & \nmc{0.4384}{0.0349} & \nmc{0.3912}{0.0237} & \nmc{0.6893}{0.0136} \\
        FFCL            & 2.4M   & \nmc{0.7848}{0.0038} & \nmc{0.8448}{0.0065} & \nmc{0.8569}{0.0051} & \nmc{0.3979}{0.0104} & \nmc{0.3732}{0.0188} & \nmc{0.6783}{0.0120} \\
        ST-Transformer  & 3.5M   & \nmc{0.7966}{0.0023} & \nmc{0.8521}{0.0026} & \nmc{0.8707}{0.0019} & \nmc{0.3984}{0.0228} & \nmc{0.3765}{0.0306} & \nmc{0.6823}{0.0190} \\
        \midrule
        BIOT            & 3.2M   & \nmc{0.7959}{0.0057} & \nmc{0.8792}{0.0023} & \nmc{0.8815}{0.0043} & \nmc{0.5281}{0.0225} & \nmc{0.5273}{0.0249} & \nmc{0.7492}{0.0082} \\
        LaBraM-Base     & 5.8M   & \nmc{0.8140}{0.0019} & \nmc{0.8965}{0.0016} & \smc{0.9022}{0.0009} & \nmc{0.6409}{0.0065} & \nmc{0.6637}{0.0093} & \nmc{0.8312}{0.0052} \\
        CBraMod         & 4.0M   & \nmc{0.7891}{0.0030} & \nmc{0.8636}{0.0063} & \nmc{0.8606}{0.0057} & \nmc{0.6671}{0.0107} & \nmc{0.6772}{0.0096} & \nmc{0.8342}{0.0064} \\
        CSBrain         & 4.9M   & \nmc{0.8172}{0.0043} & \nmc{0.9005}{0.0066} & \nmc{0.8957}{0.0046} & \smc{0.6903}{0.0059} & \tmc{0.6833}{0.0047} & \nmc{0.8333}{0.0057} \\
        REVE-Base       & 69M    & \smc{0.8315}{0.0014} & \bmc{0.9281}{0.0009} & \bmc{0.9245}{0.0013} & \tmc{0.6759}{0.0229} & \nmc{0.6783}{0.0199} & \smc{0.8451}{0.0129} \\
        CodeBrain & 15M    & \tmc{0.8294}{0.0013} & \smc{0.9100}{0.0006} & \tmc{0.9030}{0.0009} & \nmc{0.6428}{0.0062} & \smc{0.6912}{0.0101} & \tmc{0.8362}{0.0048} \\
        \modeltab       & 5.0M   & \bmc{0.8324}{0.0007} & \tmc{0.9030}{0.0018} & \nmc{0.8899}{0.0022} & \bmc{0.7210}{0.0259} & \bmc{0.7187}{0.0434} & \bmc{0.8514}{0.0229} \\
        \modeltab*      & 5.0M   & \nmc{0.8327}{0.0033} & \nmc{0.9054}{0.0027} & \nmc{0.8942}{0.0015} & \nmc{0.7236}{0.0283} & \nmc{0.7222}{0.0153} & \nmc{0.8539}{0.0078} \\
        \bottomrule
    \end{tabular}
    \label{tab:merged}
\end{table*}

\subsection{Comparison Baselines}
As listed in Table~\ref{tab:merged}, we compare our model with two major types of baselines: (i) task-specific EEG decoding models and (ii) foundation models for EEG decoding. The baselines are introduced briefly in the following.

Task-specific EEG decoding models include:
\begin{itemize}
    \item \textbf{EEGNet}~\cite{lawhern2018eegnet} is a compact, lightweight Convolutional Neural Network (CNN) specifically engineered for EEG, utilizing depthwise and separable convolutions to extract robust temporal and spatial features.
    \item \textbf{EEGConformer}~\cite{song2022conformer} integrates CNNs with self-attention mechanisms to seamlessly capture local spatio-temporal patterns alongside global, long-range temporal dependencies in EEG signals.
    \item \textbf{SPaRCNet}~\cite{jing2023development} is a deep convolutional neural network with dense residual connections, designed for expert-level classification of seizures and periodic/rhythmic patterns in clinical EEG monitoring.
    \item \textbf{ContraWR}~\cite{yang2023self} is a self-supervised representation learning framework that converts EEG signals into spectrograms and employs contrastive learning on top of a 2D ResNet backbone to learn robust feature embeddings.
    \item \textbf{FFCL}~\cite{li2022motor} is a hybrid deep learning model that effectively captures comprehensive EEG dynamics by fusing spatial representations from a CNN branch with temporal features extracted by a parallel LSTM encoder.
    \item \textbf{ST-Transformer}~\cite{song2021transformer} is a Transformer-based model where the attention mechanisms are applied to both spatial and temporal dimensions to capture the structural dependencies in EEG signals.
\end{itemize}

Foundation models for EEG decoding include:
\begin{itemize}
    \item \textbf{BIOT}~\cite{yang2023biot} is a flexible, transformer-based framework that utilizes spectral energy vectors extracted from raw signals to facilitate scalable supervised or self-supervised pre-training across diverse setups.
    \item \textbf{LaBraM-Base}~\cite{jiang2024labram} is a pioneering EEG foundation model that tokenizes continuous raw EEG patches using a vector-quantized neural spectrum tokenizer and performs self-supervised pre-training via masked neural code prediction.
    \item \textbf{CBraMod}~\cite{wang2025cbramod} is a criss-cross EEG foundation model that decomposes spatial and temporal dependencies into parallel attention mechanisms to address the spatio-temporal structure of EEG signals.
    \item \textbf{CSBrain}~\cite{zhou2025csbrain} is a cross-scale spatio-temporal EEG foundation model that employs cross-scale tokenization and structured sparse attention to capture multi-resolution neural dynamics.
    \item \textbf{REVE-Base}~\cite{el2026reve} is an adaptable EEG foundation model that utilizes a 4D positional encoding scheme and masked autoencoding to generalize robustly across arbitrary electrode layouts and recording protocols.
    \item \textbf{CodeBrain}~\cite{ma2026codebrain} is an EEG foundation model combining a decoupled tokenizer and a multi-scale attention mechanism to model both inter- and intra-patch dependencies.
\end{itemize}

\subsection{Evaluation Metrics}

Consistent with prior works~\cite{wang2025cbramod,zhou2025csbrain}, we report Balanced Accuracy, AUC-PR, and AUROC for binary classification tasks; Balanced Accuracy, Cohen's Kappa, and Weighted F1-score for multiclass classification tasks.
Balanced Accuracy is the average recall across classes, providing a class-balanced measure of classification performance. Cohen's Kappa measures agreement between predictions and ground-truth labels beyond chance, while Weighted F1-score averages per-class F1-scores weighted by class support. AUC-PR and AUROC measure the areas under the precision-recall and receiver operating characteristic curves, respectively, with AUC-PR being particularly informative for imbalanced datasets and AUROC assessing the model's ability to distinguish positive from negative samples.
We use Balanced Accuracy as the monitor score to select the best model for evaluation.

\begin{table}[t]
    \centering
    \small
    \setlength{\tabcolsep}{0.3em}
    \caption{Ablation results reported in Balanced Accuracy (mean $\pm$ std). Note that the four scale-related setups form a $2\times2$ factorial decomposition over input-side and objective-side scale modeling. The \textbf{best} results are marked.}
    \begin{tabular}{lcc}
        \toprule
        \textbf{Setup} & \textbf{PhysioNet-MI} & \textbf{SEED-V} \\
        \midrule
        w/o Pre-training    & \nmc{0.6180}{0.0047} & \nmc{0.3957}{0.0048} \\
        w/o Scale-Aware PEs & \nmc{0.6458}{0.0081} & \nmc{0.4111}{0.0088} \\
        \midrule
        w/o Multi-Scale     & \nmc{0.6358}{0.0079} & \nmc{0.4100}{0.0050} \\
        w/o MS Tokenization & \nmc{0.6518}{0.0067} & \nmc{0.4054}{0.0095} \\
        w/o Scale-Specific Recon & \nmc{0.6402}{0.0054} & \nmc{0.3969}{0.0052} \\
        Full                & \bmc{0.6524}{0.0111} & \bmc{0.4116}{0.0056} \\
        \midrule
        \textbf{Setup} & \textbf{HMC} & \textbf{MentalArithmetic}\\
        \midrule
        w/o Pre-training    & \nmc{0.7277}{0.0047} & \nmc{0.5792}{0.0193} \\
        w/o Scale-Aware PEs & \nmc{0.7504}{0.0091} & \nmc{0.7229}{0.0344} \\
        \midrule
        w/o Multi-Scale     & \nmc{0.7471}{0.0064} & \nmc{0.7292}{0.0834} \\
        w/o MS Tokenization & \nmc{0.7445}{0.0071} & \nmc{0.6764}{0.0307} \\
        w/o Scale-Specific Recon & \nmc{0.7532}{0.0057} & \nmc{0.6882}{0.0498} \\
        Full                & \bmc{0.7540}{0.0098} & \bmc{0.7771}{0.0429} \\
        \midrule
        \textbf{Setup} & \textbf{TUAB} & \textbf{TUEV} \\
        \midrule
        w/o Pre-training    & \nmc{0.7902}{0.0061} & \nmc{0.6736}{0.0354} \\
        w/o Scale-Aware PEs & \nmc{0.8320}{0.0038} & \bmc{0.7312}{0.0145} \\
        \midrule
        w/o Multi-Scale     & \bmc{0.8337}{0.0031} & \nmc{0.6986}{0.0212} \\
        w/o MS Tokenization & \nmc{0.8301}{0.0042} & \nmc{0.6932}{0.0169} \\
        w/o Scale-Specific Recon & \nmc{0.8324}{0.0029} & \nmc{0.6945}{0.0221} \\
        Full                & \nmc{0.8324}{0.0007} & \nmc{0.7210}{0.0259} \\
        \bottomrule
    \end{tabular}
    \label{tab:ablation}
\end{table}

\subsection{Performance Comparison}
We compare ScaleBrain with 12 representative baselines, including 6 task-specific models and 6 EEG foundation models, across 6 BCI classification datasets. 
The results are presented in Table~\ref{tab:merged}. For ScaleBrain, we report the mean and standard deviation over five random seeds, while baseline results are taken from the corresponding works when available.

Overall, EEG foundation models provide stronger baselines than task-specific models on most datasets, suggesting the benefit of large-scale EEG pre-training for transferable representation learning. Among foundation models, ScaleBrain achieves the best Balanced Accuracy on five of the six datasets and ranks third on SEED-V behind CSBrain and CodeBrain. The largest gain appears on TUEV, where ScaleBrain reaches 0.7210 and improves over CSBrain by 3.07 percentage points. It also obtains strong results on MentalArithmetic and HMC, with balanced accuracy of 0.7771 and 0.7540, respectively.
The improvements over existing multi-scale foundation models further suggest that coordinating multi-scale representations across the pre-training pipeline can provide benefits beyond simply increasing scale diversity.
As TUAB and TUEV are subsets of the TUEG corpus, we additionally exclude recordings associated with TUAB and TUEV from TUEG pre-training to examine the effect of potential pre-training exposure. The resulting ``ScaleBrain*'' achieves comparable or slightly better performance than the full-TUEG setting, indicating that the observed performance is not directly attributable to exposure of downstream recordings.

ScaleBrain also shows favorable parameter efficiency. With 5.0M parameters, it outperforms REVE-Base across all six datasets, despite REVE-Base using 69M parameters.
These results indicate that scale-consistent multi-scale pre-training can improve EEG representation learning without relying on a substantially larger model size.

\begin{figure*}[ht]
    \centering
    \includegraphics[width=0.96\linewidth]{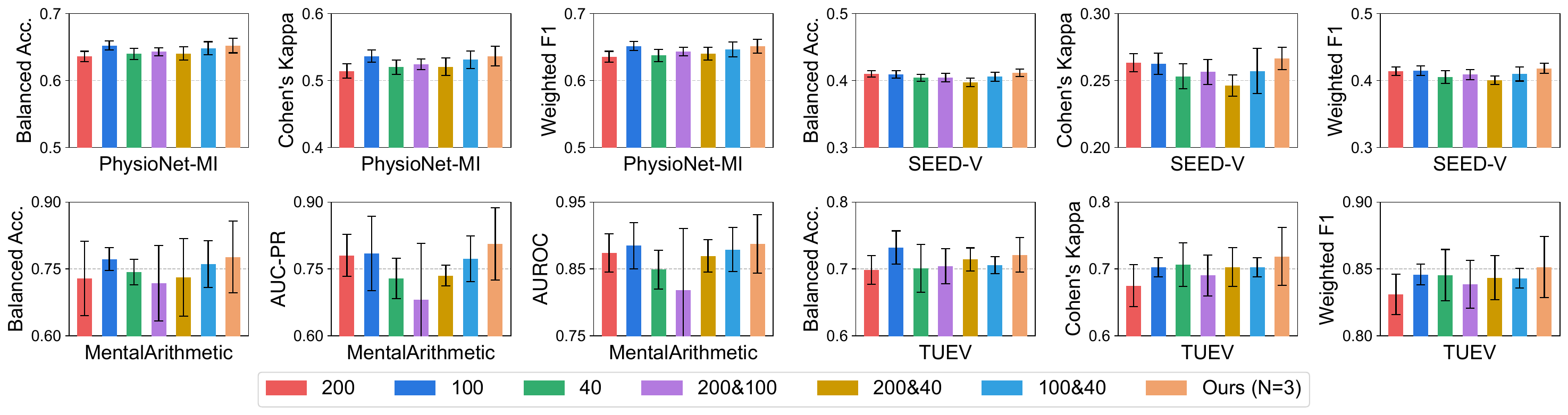}
    \caption{Analysis of the necessity of scales. Acc.: Accuracy.}
    \label{fig:analysis_scales}
\end{figure*}

\begin{figure*}[ht]
    \centering
    \includegraphics[width=\linewidth]{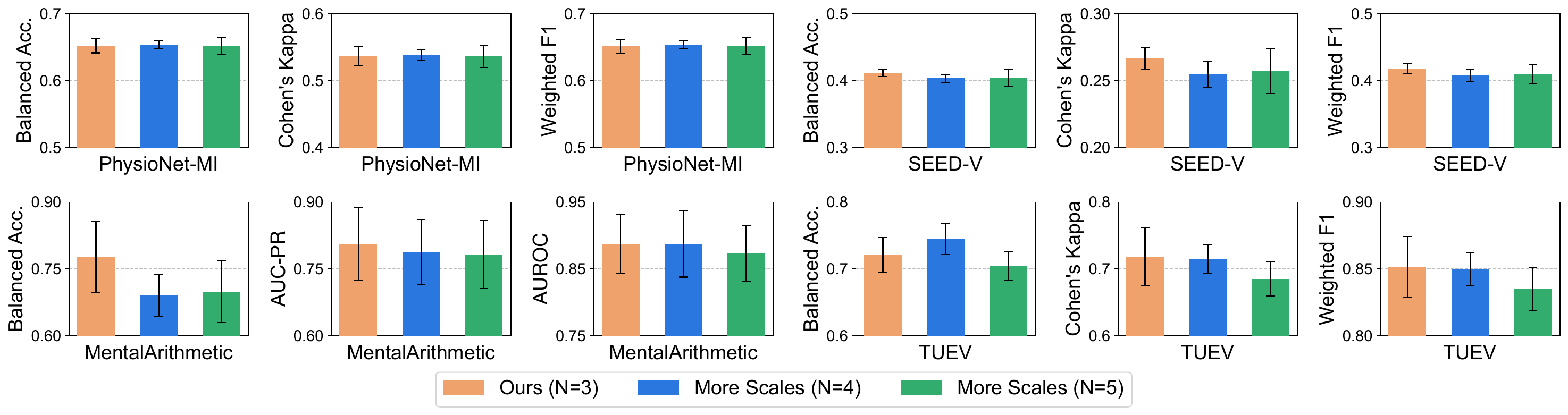}
    \caption{Analysis of the effect of additional scales. Acc.: Accuracy.}
    \label{fig:analysis_more}
\end{figure*}

\begin{figure*}[ht]
    \centering
    \includegraphics[width=\linewidth]{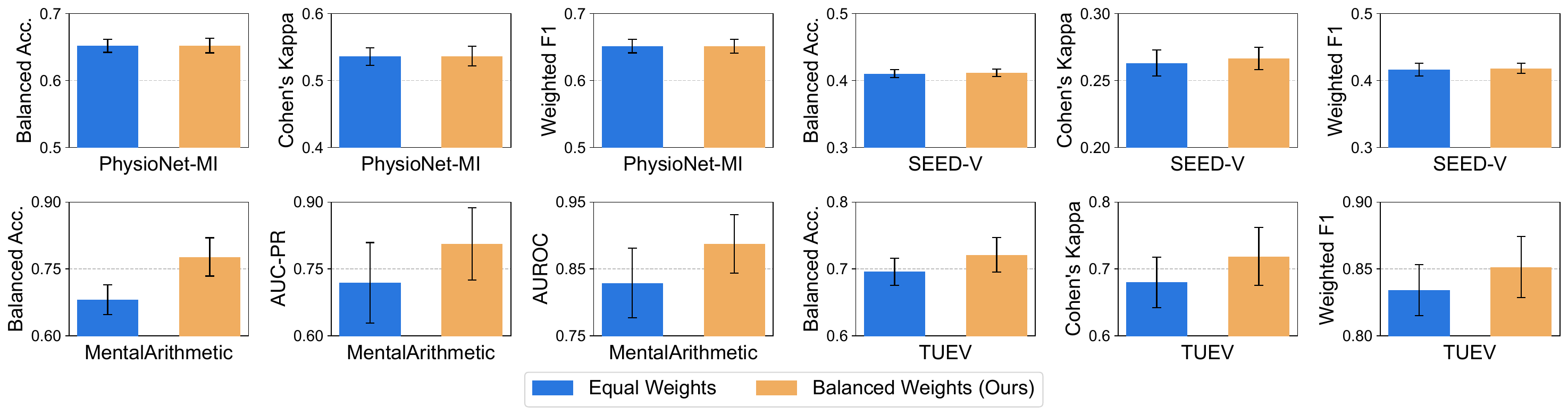}
    \caption{Analysis of the objective weights. Acc.: Accuracy.}
    \label{fig:analysis_weights}
\end{figure*}

\begin{figure*}[ht]
    \centering
    \includegraphics[width=0.96\linewidth]{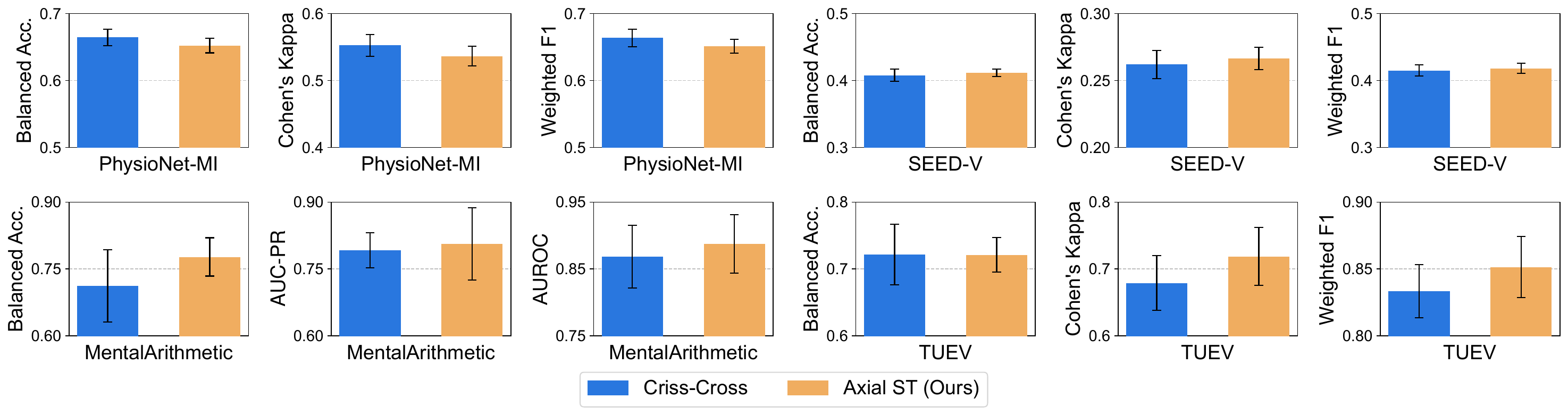}
    \caption{Analysis of the attention mechanism. Acc.: Accuracy.}
    \label{fig:analysis_attn}
\end{figure*}

\subsection{Ablation Study}
To evaluate the main components of ScaleBrain and examine the role of scale consistency during pre-training, we conduct ablation experiments in Table~\ref{tab:ablation}.

\pars{Effectiveness of Pre-Training.}
To assess self-supervised pre-training, we compare the full model with a variant trained from scratch (``w/o Pre-training''). This variant keeps the same architecture but is trained directly on each downstream dataset with a classification head. Removing pre-training consistently degrades performance across all 6 datasets, with the largest drop on MentalArithmetic, confirming the importance of large-scale self-supervised pre-training for EEG representation learning.

\pars{Effectiveness of Scale-Aware PEs.}
To assess scale-aware PEs, we compare the full model with a variant using shared PEs across scales (``w/o Scale-Aware PEs''), where the same spatial and temporal PEs are applied after the gated scale fusion. Scale-aware PEs improve performance on 5 of the 6 datasets. The only exception is TUEV, where shared PEs achieve 0.7312 compared with 0.7210 for the full model, suggesting that explicit scale-aware positional alignment brings generally useful but task-dependent benefits. Nevertheless, the scale-aware formulation provides a generally effective way to preserve scale identity while maintaining cross-scale temporal correspondence.

\pars{Effect of Input- and Objective-Side Multi-Scale Modeling.}
The four scale-related configurations in Table~\ref{tab:ablation} form a $2\times2$ factorial design that separates input-side multi-scale representation from objective-side scale-specific supervision, allowing us to examine their interaction. Specifically, ``w/o Multi-Scale'' uses a single-scale input representation and a single-scale reconstruction objective; ``w/o MS Tokenization'' uses single-scale tokenization with scale-specific reconstruction; ``w/o Scale-Specific Recon'' uses multi-scale tokenization with a single-scale reconstruction objective; and the full model uses both multi-scale tokenization and scale-specific reconstruction.
This design allows us to examine the effects of input-side and objective-side multi-scale modeling as well as their interaction.

Across the six datasets, the two factors exhibit complementary but task-dependent effects. Notably, the effect of scale-specific reconstruction depends strongly on the presence of multi-scale tokenization. For example, on MentalArithmetic, scale-specific reconstruction decreases balanced accuracy from 0.7292 to 0.6764 when the input remains single-scale, but increases it from 0.6882 to 0.7771 when multi-scale tokenization is used. This interaction suggests that objective-side scale-specific supervision is most effective when coupled with multi-scale input representations.
Overall, the factorial comparison supports the central design of ScaleBrain: multi-scale input representation and scale-specific supervision should be jointly coordinated rather than introduced independently.

\subsection{Further Analyses}
To further analyze ScaleBrain, we study the scale selection, the attention mechanism, and the objective weights.

\pars{Analysis of the Necessity of Scales.}
To validate the necessity of coarse, medium, and fine scales ($s_1=200, s_2=100, s_3=40$), we compare the full ScaleBrain model against three single-scale configurations and three double-scale combinations on four representative datasets. As illustrated in Fig.~\ref{fig:analysis_scales}, the multi-scale combinations generally outperform their single-scale counterparts. Furthermore, the three-scale configuration (``Ours'') consistently yields the best or highly competitive results across Balanced Accuracy, Cohen's Kappa, and Weighted F1. This suggests that coarse-to-fine temporal scales provide complementary information for EEG representation learning.

\pars{Analysis of the Effect of Additional Scales.}
To investigate whether incorporating more temporal scales beyond the default $N=3$ can further enhance representation capacity, we conduct experiments with larger scale numbers, specifically $N=4$ and $N=5$. For $N=4$, we add a new scale $s_4=20$, and for $N=5$, we further add $s_5=10$. We evaluate these variants across four representative datasets.
As illustrated in Fig.~\ref{fig:analysis_more}, the results consistently show that the default $N=3$ configuration yields the most favorable performance among the evaluated configurations. On PhysioNet-MI, SEED-V, and MentalArithmetic, increasing the scale number to $N=4$ or $N=5$ leads to varying degrees of performance degradation and sometimes introduces larger variances. On the TUEV dataset, while the $N=5$ variant achieves comparable results to the baseline, it does not provide a meaningful improvement. This phenomenon suggests that simply stacking more temporal scales does not necessarily improve representation learning. Instead, an excessive number of scales may introduce redundant representations and increase optimization difficulty. Therefore, the coarse, medium, and fine three-scale partition ($N=3$) provides a favorable trade-off between temporal coverage, representation diversity, and optimization complexity.

\pars{Analysis of the Objective Weights.}
To validate the effectiveness of the scale-balanced design of the objective weights, \textit{i.e.}, $\lambda_1=0.5$, $\lambda_2=1.0$, $\lambda_3=2.0$ (``Balanced Weights''), we conduct pre-training with ``Equal Weights'' ($\lambda_1=\lambda_2=\lambda_3=1.0$) and evaluate the model on four representative datasets.
As illustrated in Fig.~\ref{fig:analysis_weights}, the proposed scale-balanced weighting scheme consistently outperforms the equal-weight baseline across all four datasets and all evaluated metrics. Notably, the performance improvements are particularly pronounced on the MentalArithmetic and TUEV datasets.
One possible explanation is that the reconstruction targets at different scales exhibit different statistical distributions and energy characteristics. Under equal weighting, these differences may cause the optimization to favor some scales over others.
We therefore assign progressively larger weights to finer temporal resolutions to reduce optimization bias toward coarse-scale structure and balance the contribution of different temporal scales. These results provide empirical support for our balanced objective weighting design.

\pars{Analysis of the Attention Mechanism.}
We compare Axial ST attention with the criss-cross attention used in CBraMod~\cite{wang2025cbramod}. For a fair comparison, the criss-cross variant uses 12 layers to maintain a similar parameter budget. Fig.~\ref{fig:analysis_attn} reports the results on four datasets, where Axial ST achieves competitive or superior performance across the majority of evaluated datasets and metrics, although criss-cross attention performs better on some individual settings. These results suggest that decoupling spatial and temporal attention provides a competitive alternative for contextualizing the fused multi-scale representation, while neither attention design dominates uniformly across all downstream tasks.

\section{Conclusion}
In this paper, we propose \textbf{ScaleBrain}, a scale-consistent pre-training framework for EEG foundation models that addresses the limitation of current multi-scale pre-training.
By incorporating multi-scale input tokenization and scale-specific reconstruction targets, ScaleBrain explicitly models EEG representations at different temporal resolutions from both the input and optimization objective sides. Scale-aware spatio-temporal position encodings preserve spatial and temporal correspondence across scales, while gated scale fusion and axial spatio-temporal attention integrate and contextualize the resulting multi-scale representations.
Extensive evaluations across six BCI benchmarks demonstrate the effectiveness of the proposed framework. ScaleBrain achieves the best mean Balanced Accuracy on five datasets and ranks third on the remaining one, while using only 5.0M parameters.
These results demonstrate that effective multi-scale pre-training requires coordinated representation and supervision across temporal scales.
Moving forward, we plan to extend this scale-consistent approach to other neuroimaging modalities such as magnetoencephalography (MEG), and investigate more efficient attention mechanisms to improve the practicality of large-scale EEG foundation models.


\nolinenumbers

\bibliographystyle{IEEEtran}
\bibliography{references}

\vfill

\end{document}